\documentclass[sigconf,balance=False]{acmart}

\AtBeginDocument{%
  \providecommand\BibTeX{{%
    \normalfont B\kern-0.5em{\scshape i\kern-0.25em b}\kern-0.8em\TeX}}}

\copyrightyear{2023} 
\acmYear{2023} 
\setcopyright{acmlicensed}\acmConference[GECCO '23]{Genetic and Evolutionary Computation Conference}{July 15--19, 2023}{Lisbon, Portugal}
\acmBooktitle{Genetic and Evolutionary Computation Conference (GECCO '23), July 15--19, 2023, Lisbon, Portugal}
\acmPrice{15.00}
\acmDOI{10.1145/3583131.3590469}
\acmISBN{979-8-4007-0119-1/23/07}

\acmConference[GECCO '23]{GECCO '23: Genetic and Evolutionary Computation Conference}{July 15--19, 2023}{Lisbon, Portugal}

\usepackage{hyperref}

\begin{document}

\title{Selection for short-term empowerment accelerates the evolution of homeostatic neural cellular automata.}

\author{Caitlin Grasso}
\email{Caitlin.Grasso@uvm.edu}
\affiliation{%
  \institution{University of Vermont}
  \city{Burlington}
  \state{Vermont}
  \country{USA}
}

\author{Josh Bongard}
\email{Josh.Bongard@uvm.edu}
\affiliation{%
  \institution{University of Vermont}
  \city{Burlington}
  \state{Vermont}
  \country{USA}
}

\begin{abstract}

Empowerment---a  domain independent, information-theoretic metric---has previously been shown to assist in the evolutionary search for neural cellular automata (NCA) capable of homeostasis when employed as a fitness function \cite{klyubin2005empowerment, grasso2022empowered}. In our previous study, we successfully extended empowerment, defined as maximum time-lagged mutual information between agents' actions and future sensations, to a distributed sensorimotor system embodied as an NCA. However, the time-delay between actions and their corresponding sensations was arbitrarily chosen. Here, we expand upon previous work by exploring how the time scale at which empowerment operates impacts its efficacy as an auxiliary objective to accelerate the discovery of homeostatic NCAs. We show that shorter time delays result in marked improvements over empowerment with longer delays, when compared to evolutionary selecting only for homeostasis. Moreover, we evaluate stability and adaptability of evolved NCAs, both hallmarks of living systems that are of interest to replicate in artificial ones. We find that short-term empowered NCA are more stable and are capable of generalizing better to unseen homeostatic challenges. Taken together, these findings motivate the use of empowerment during the evolution of other artifacts, and suggest how it should be incorporated to accelerate evolution of desired behaviors for them.\footnote{Source code and videos for the experiments in this paper can be found at: \url{https://github.com/caitlingrasso/empowered-nca-II}.}

\end{abstract}




\maketitle

\section{Introduction}

Many collective systems of interest are comprised of large numbers of components that operate according to simple, local rules to coordinate system-level behavior. Such systems are increasingly pertinent in the fields of robotics and artificial intelligence as they confer desired characteristics such as emergence, robustness, and flexibility. This is seen at many different scales in biology, and in particular in morphogenesis, where a single cell rapidly proliferates, and its many progeny self-organize into a stable target shape. Understanding the mechanisms by which collective systems communicate information to achieve such global coordination could provide key insights for self-assembling robots \cite{tuci2006cooperation, klavins2007programmable, rubenstein2014programmable}, shape-changing robots \cite{yu2008morpho, kim2018printing, shah2021shape, baines2022multi}, and regenerative medicine systems \cite{mitchell2016engineering, kuchling2020morphogenesis, levin2021bioelectric}. Cellular automata are particularly useful models of collective growth and behavior due to their simplicity, and local and emergent nature. Finding general principles of morphogenetic and homeostatic systems via models such as CAs could suggest hypotheses regarding the biological systems they model as well as design guidelines for autonomous machines that can adapt morphologically to internal and external changes, for example. 


Often, the local rulesets upon which CA operate are evolved such that the CA exhibits some global behaviors or characteristics. The evolved CA can then be examined in an attempt to understand their complex dynamics. The fitness functions in these experiments are typically specific to the CA, however, in recent years, the concept of universal objective functions has emerged in the fields of robotics and artificial life, where the traditional task-specific fitness function is replaced with a task-independent function that selects for generally useful or interesting behaviors \cite{eysenbach2018diversity, lehman2011evolving, pathak2017curiosity, campero2020learning, chan2018lenia, turner2019optimal}. One such class of these universal functions makes use of information-theoretic measures to capture certain behavioral dynamics \cite{ay2008predictive, pitis2020maximum, der2008predictive}. Of particular interest is empowerment \cite{salge2014empowerment}, which attempts to measure the amount of control an agent has over its environment by the channel capacity of the agent's sensorimotor channel. In other words, empowerment is the maximum time-lagged mutual information between an agent's actions in the environment (i.e. channel) and future sensor states. 

In previous work, we showed that empowerment could be expanded to a distributed, multi-agent system embodied by a neural cellular automata (NCA) where each cell is considered an agent and its neighboring cells constitute its environment. The states of the cells in the NCA are updated via a neural network. Cells can sense their neighbors and sensor states are defined as inputs to network. Their actions are defined as outputs from the network, which determine how they update for the next time step. Thus, empowerment as a evolutionary objective selects for NCA in which unique cellular actions elicit unique sensations. We hypothesized this behavior would be helpful for a collective task such as morphogenesis and homeostasis when applied as an additional objective in multiobjective search as it requires cells to coordinate with neighbors to produce these unique sensor/action pairs. We found a synergism between the objective that selected for morphogenesis and homeostasis and the empowerment objective: the addition of empowerment as an auxiliary objective helped evolve NCA that were better at matching and maintaining a target shape, compared to evolving for shape-matching alone. 

In that work the time horizon (delay between actions and future sensations) used for calculating empowerment was chosen arbitrarily. However, the time scale at which information is communicated through a system is an important parameter. Time scale can be investigated by altering the lag between the two variables of interest, in this case, between actions and sensor states. The importance of time scale is known from, for instance,  neuroscience, where effective connectivity networks are often inferred from brain signals to understand behavior \cite{varley2023information}. The time-lag at which effective connectivity measures are computed can be key to realizing certain behaviors. Effective connectivity, and the time scale at which it is established, also appears to play a role in biochemical networks \cite{manicka2022effective}. In the empowerment literature, time horizon is typically a constant hyperparameter set prior to experimentation \cite{capdepuy2007maximization, capdepuy2012perception, clements2017empowerment}, although some studies suggest is should be set depending on the particular task at hand \cite{guckelsberger2014effects, johansson2009prediction}.

Here, we explore the time horizon in the computation of multi-agent empowerment and assess how or whether it affects empowerment's ability to facilitate the evolution of morphogenesis and homeostasis in NCA. As reported in the results, we find that evolving for coordination on short time scales produce better results. We show that this result holds for (1) NCA with more cells; (2) for other morphogenetic and homeostatic tasks (in the form of growing to and holding different target shapes); and (3) that such NCA can maintain homeostasis for longer periods (they hold their shapes better beyond the time interval for which they were evolved to do so). We also find that NCA evolved with short-term empowerment can be more easily adapted by evolution to other morphogenetic and homeostatic tasks. This is roughly akin to pre-training in deep learning models \cite{erhan2010does, hendrycks2019using, zoph2020rethinking}. In the discussion section we hypothesize why shorter history lengths produce better results and what types of action patterns empowerment is selecting for. We conclude by mapping out future research directions. Our findings further recommend short-term empowerment as a useful domain-independent auxiliary objective function particularly in the development of machines wishing to maintain form and function in the face of surprise. 



\section{Methods}

We employ Age Fitness Pareto Optimization (AFPO) \cite{schmidt2010age} to perform several evolutionary bi- or tribjective optimization experiments. The objectives include some combination of the following:

\begin{enumerate}
    \item \textit{Age.} This is the age, in generations, of the species to which the current genome belongs. By minimizing age as an objective, the AFPO algorithm is better able to maintain diversity in the population. Thus, it is always included as an objective in all evolutionary runs.
    \item \textit{Loss.} Morphogenesis and homeostatis are simplified to the task of reaching and maintaing a specified target shape. The ability of the NCA to achieve this is quantified by the loss between the final shape of the NCA after simulation and the target shape.
    \item \textit{Empowerment.} A task-independent measure which selects for coordination between cells in the NCA at some time scale as determined by the time horizon parameter. 
\end{enumerate}

We test which combinations of these objectives, and in particular which values of time horizon for the empowerment objective, help the most in finding NCA capable of matching the target shape. A brief overview of the NCA and experimental framework are provided. For more detail please refer to \cite{grasso2022empowered}.

\subsection{NCA Model}

\begin{figure}[htbp]
  \centering
  \includegraphics[width=\linewidth]{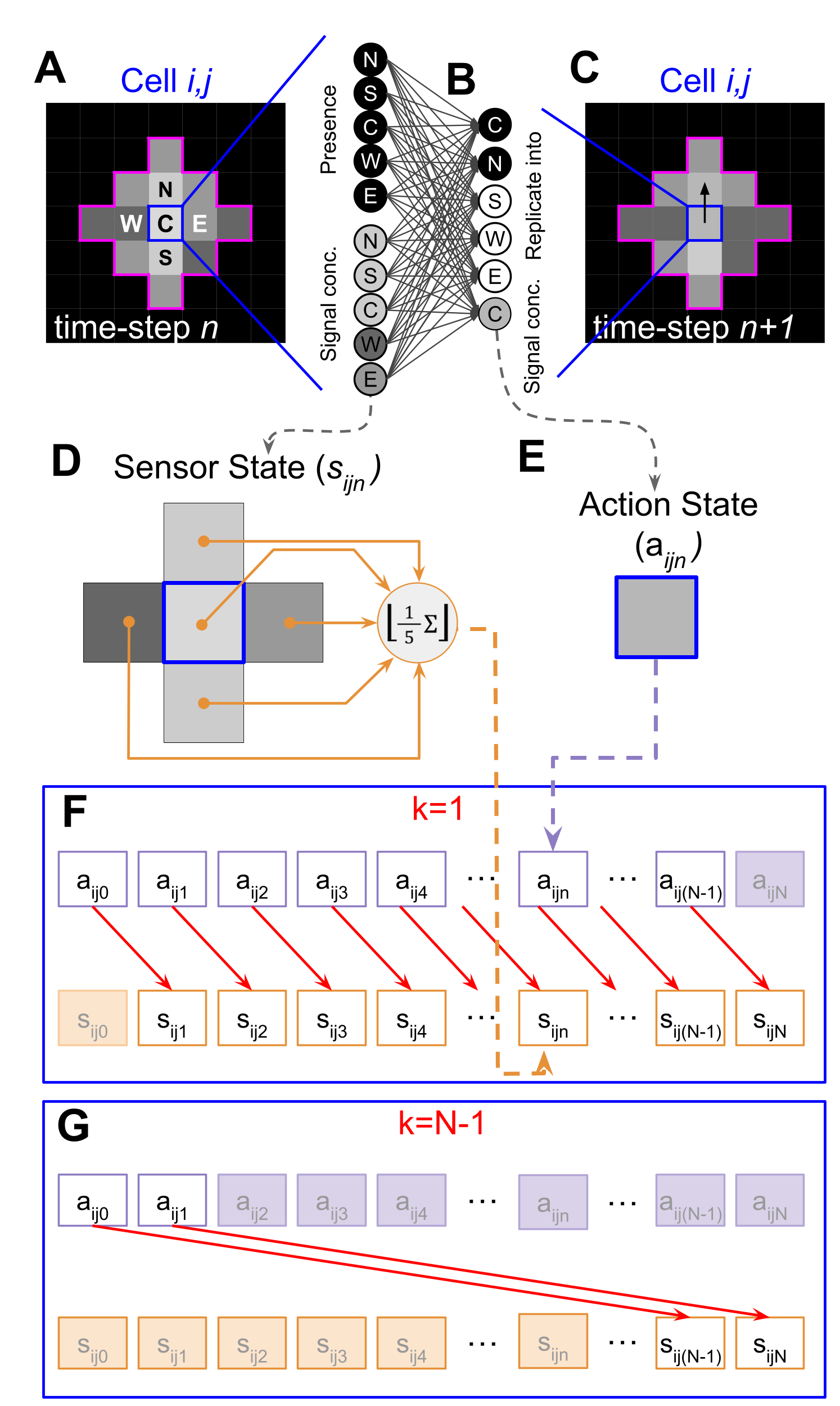}
  \caption{Construction of sensorimotor time series for empowerment calculation with different time horizons, $k$. At each time-step $n$ during the NCA simulation (A) the neural network (B) is executed for each live cell $i,j$ to produce the state of the NCA at the following time-step $n+1$ (C). Neural network inputs and outputs define the sensor state $s_{ijn}$ (D) and action state $a_{ijn}$ (E), respectively, based on the signaling channel associated with a given cell and time-step. These values are collected over the total length of simulation, $N$, to construct action and sensor time series for cell $i,j$ (F,G). The sensor time series are shifted according to the time horizon $k$ to produce a set of sensor/action pairs (unshaded boxes). Shaded boxes are not included in the empowerment calculation due to the value of the time horizon which can either be short (F), long (G) or somewhere in between. }
  \Description{Images of a toy cellular automata update displaying the neural network inputs and outputs. Arrows from neural network inputs and outputs point to cartoons describing the sensor and action states of the cell which are then pointed to in the sensor/action time series. Two different sensor, action time series are shown with different values of k represented by red arrows that point from an action value to it's corresponding sensor value depending on the lag. This is shown for k=1 and k=N-1.}
  \label{fig:methods}
\end{figure}

The cellular automata (CA) consists of two channels: a binary live/dead channel, and a signaling channel with values ranging from 0-255 (Fig. \ref{fig:methods}A). Discrete values of signal concentration were employed in order to simplify the empowerment calculation which traditionally operates on discrete states. Although the signaling channel has no predefined meaning, it is provided to allow for an additional mechanism of communication between cells beyond just whether neighbors are alive or dead. Each live cell in the CA is governed by an identical feed-forward neural network with no hidden layers (Fig. \ref{fig:methods}B). 

The ten inputs to the network include the binary states of the four cells in its Von Neumann neighborhood and itself, and the signaling concentrations of the four cells in its Von Neumann neighborhood and itself. The network outputs an updated binary state and signal concentration for the cell as well as four bits that determine whether or not the cell should replicate its binary state and signal concentration into its neighbors in the respective directions (Fig. \ref{fig:methods}C). Replication occurs regardless of whether the target cells is alive or dead by overriding the binary state and signal of the target cell. These binary replication outputs allow for growth or death of contiguous patches of cells. Outputs of the network pass through a sigmoid activation function and are then binarized or scaled depending on the output type. For the binary outputs, if the activation is greater than $\epsilon$, the output is $1$, otherwise it is set to $0$. For the signal concentration node, if the output is greater than $\epsilon$, the value is scaled between $[127,255]$ and binned to an integer, otherwise it is set to $0$. In this way, the NCA is heavily biased towards growth and signaling. As in biological morphogenesis, proliferation and signaling are more common than cell death or quiescence making controlled growth a challenging task. Updates are performed in sequential order from the top left to bottom right of the NCA grid.

The initial condition of the NCA is always a single cell in the center of an $M \times M$ 2D grid with zero signal concentration. During evolution the NCA is evaluated by performing $N$ updates where neural networks in each live cell are executed sequentially and synchronously to determine the new state of the CA. After $N$ iterations, the simulation terminates and loss and/or empowerment are computed.

\subsection{Empowerment and Time Horizon}

The empowerment objective attempts to select for NCA that  exert control over their environment, which in this domain is the cells in their neighborhood.  As traditionally defined for a sensorimotor agent, empowerment $\mathfrak{E}$ is the maximum mutual information between the agent's action states ($A$) and future sensor states ($S$) as described in Eqn. \ref{eqn:empowerment}:

\begin{equation}
    \mathfrak{E}(k) = I(A_0^{k}, S_{N-k}^{N})
    \label{eqn:empowerment}
\end{equation}

where $I$ is mutual information and $k$ is the time-lag between an action and its corresponding sensor state. This calculation is expanded to cellular automata by considering each cell as an agent with a set of actions and sensors. Cellular sensors and actions are computed from the input and output values of the neural network, respectively. Only the inputs and outputs pertaining to signal detection and signal emission are employed for this calculation. In other words, cells can be thought to have four exteroceptive sensors that sense the signals of the cells' immediate neighbors, and one interoceptive sensor that senses its own signal concentration. The sensor state of cell $i,j$ at time-step $n$, $s_{ijn}$, is defined as the mean of these five sensed values, binned back into an integer in $[0,255]$ (Fig. \ref{fig:methods}D). An action is defined as $a_{ijn} \in [0,255]$, the new signal concentration that cell $i,j$ assigns to itself at time step $n$. This concentration is made available for cells in its neighborhood to sense at the following time-step. By combining all actions and sensors for every cell at every time-step, we construct the two motor and sensory time series data sets

\begin{displaymath}
  A_0^{k} = \{ 
  a_{000},
  \cdots, 
  a_{MM0},
  a_{001},
  \cdots,
  a_{MMk}\}\\
\end{displaymath}

and

\begin{displaymath}
    S_{N-k}^{N} = \{ 
    s_{00(N-k)},
    \cdots, 
    s_{MM(N-k)},
    s_{00(N-k+1)},
    \cdots,
    s_{MMN}\}
\end{displaymath}

Mutual information is then computed on these time series to arrive at a multi-agent form of empowerment

\begin{equation}
    \mathfrak{E}(k) = I(A_0^{k}, S_{N-k}^{N}) = -\sum_{\substack{a_{ijn} \in A_0^{k}, \\ s_{ijn} \in S_{N-k}^{N}}} p(a_{ijn},s_{ijn}) \log_2 \frac{p(a_{ijn},s_{ijn}) }{p(a_{ijn})p(s_{ijn})}
    \label{eqn:empowerment_CA}
\end{equation}
where $k$ denotes the time horizon over which empowerment is computed.

Defining sensors and actions solely based on the signaling channel in this empowerment calculation allows cells to communicate with one another, and possibly coordinate their collective action, without having to switch their live/dead status. No empowerment within an NCA can result if cells produce complete noise in their signaling channel, or produce the same amount of signal at every time-step. In either case, the NCA is not utilizing the signaling channel in a manner that may aid in its primary developmental task. Empowered NCA, on the other hand, exhibit signaling dynamics somewhere between these two extremes. The behavior captured by the empowerment calculation is impacted by the choice of the time horizon hyperparameter, $k$. With a short time horizon, cells' actions reliably predict their sensor states in the near future (Fig. \ref{fig:methods}F). Long time horizons select for reliable prediction of cells' actions and their sensor states in the distant future (Fig. \ref{fig:methods}G). The time scale at which cells should coordinate their signals to best facilitate the task of developing and maintaining a target shape is explored in the following experiments. 

\begin{figure*}[t]
  \centering
  \includegraphics[width=\textwidth]{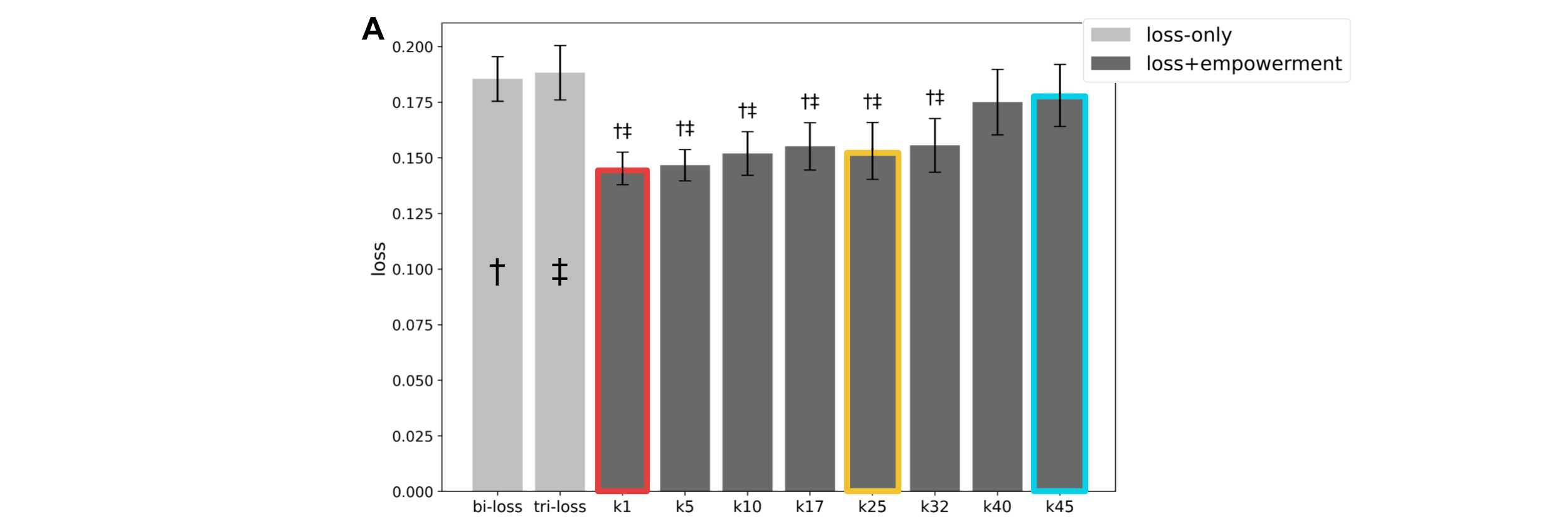}
  \includegraphics[width=\textwidth]{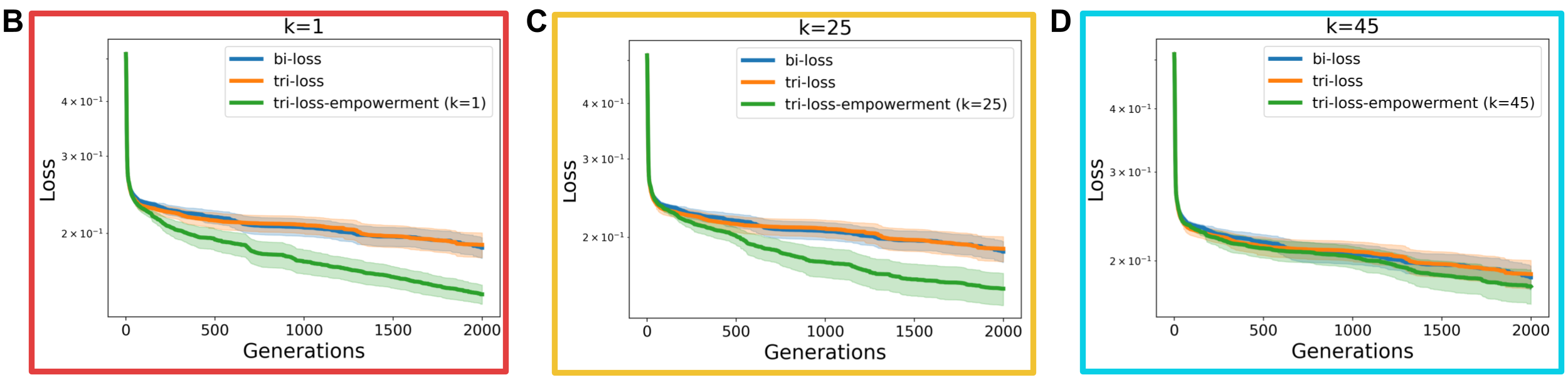}
  \caption{(A) Loss of the best NCA at the end of 2000 generations averaged over all replicates (with 95\% confidence intervals) for 10 different evolutionary variants. $\dagger$ indicates a significant difference from the bi-loss control and $\ddagger$ indicates a significant different from the tri-loss control with a Bonferroni-corrected significance level of $p<0.0031$. (B) Average loss (with 95\% confidence intervals) over evolutionary time for both controls and empowerment runs computed with a time-horizon of $k=1$. The tri-loss-empowerment curve (green) is significantly lower that the loss-only control curves (blue and orange). This gap lessens with an increase in time-horizon as seen with $k=25$ (B) and $k=45$ (C).}
  \Description{Bar plot of loss at the end of evolution across various evolutionary trials where the shorter time horizon empowerment runs have significantly lower loss than the loss-only controls. Below the bar plot are three plots of loss curves for a short, medium, and long time horizon.}
  \label{result:shortk}
\end{figure*}

\subsection{Evolutionary Experiments}

The primary objective for the NCA is to grow from a single cell into a pre-specified  target shape $T$, and hold that shape. To evaluate NCA on this task, we employ a loss measure, $\mathfrak{L}$, formally defined as the L2 loss between the NCA grid at iteration $n$ and the target averaged over all CA iterations from $n_0$ to $n_1$ (Eqn. \ref{eqn:error}). Averaging over every time-step during development rather than just the final state applies selection pressure for NCA that are able to reach the target shape more quickly and then maintain it.  

\begin{equation}
    \mathfrak{L}(n_0, n_1) = \frac{1}{n_1-n_0} \sum_{n \in [n_0, n_1]} \frac{\sum_{i,j \in M} (C_{ijn}-T_{ij})^2}{M^2}
    \label{eqn:error}
\end{equation}
where $C_{ijn}$ denotes whether or not cell i,j is alive at time n
and $T_{ij}$ denotes whether or not that cell lies within the target shape.

In previous work, we found that performing evolution with empowerment as a third objective in tri-objective search in addition to age and loss (tri-loss-empowerment) with a time horizon of $k=25$ aids in the evolution of NCA that can grow to and maintain a target shape compared to evolving for age and loss alone (bi-loss). However, we did not explore other values for $k$. Here, we test empowerment as the third objective with various values of $k$. Table \ref{tab:treatments} displays the objectives for the four evolutionary treatments. The tri-loss variant is included to control for the potential that tri-loss-empowerment variants may outperform the bi-loss variant not due to empowerment, but due to a difference in the number of objectives. The tri-loss-empowerment variant was further subdivided into eight variants with  $k \in [1,5,10,17,25,32,40,45]$. For each of the $2+8=10$ variants, 35 independent runs with differing initial random populations were performed for 2000 generations and with a population size of 800. During evaluation, each NCA was simulated for $N=50$ time-steps on an $25 \times 25$ grid ($M=25$) and compared to a target square shape of size $12 \times 12$ centered around the seed cell.

\begin{table}[htbp]
    \centering
    \begin{tabular}{|c|c|c|c|}
    \hline
     \textbf{Treatment} &  \textbf{Obj. 1} & \textbf{Obj. 2} & \textbf{Obj. 3}\\
     \hline
     Bi-loss & Age & $\mathfrak{L}(0,N)$ & - \\
     \hline
     Tri-loss-empowerment & Age & $\mathfrak{L}(0,N)$ & $\mathfrak{E}(k)$ \\
     \hline
     Tri-loss & Age & $\mathfrak{L}(0,N/2)$ & $\mathfrak{L}(N/2,N)$  \\
     \hline
    \end{tabular}
    \caption{Objectives for the three algorithm variants.}
    \label{tab:treatments}
\end{table}

\section{Results}

To assess the impact of varying the time horizon $k$ in the tri-loss-empowerment trials, we evaluated the best NCA resulting from each of the evolutionary trials based on their ability to achieve the primary task of morphogenesis as well as their homeostatic capabilities.

\subsection{Morphogenesis}

The performance of the tri-loss-empowerment trials was compared to the performances of the two loss-only controls (bi-loss $\dagger$ and tri-loss $\ddagger$) in Fig. \ref{result:shortk}A. As time horizon $k$ decreases, a decrease in loss is observed which becomes increasingly significantly lower than that of both loss-only controls with $p<0.0031$ (using a Wilcoxon Rank-Sum test and Bonferroni-corrected significance level of $\alpha=0.05/16=0.0031$, denoted by the $\dagger$ and $\ddagger$ symbols above their respective bars). Three of the tri-loss-empowerment variants with a short, medium, and long time horizon are further unpacked in Figures \ref{result:shortk}B, \ref{result:shortk}C, and \ref{result:shortk}D. The loss curves for the two loss-only controls are repeated across these panels as changing $k$ does not affect them. The green curves which represent runs that incorporate empowerment as an additional objective display notable differences across time horizons. For the shortest time horizon, $k=1$, the tri-empowerment-loss curve (green) is significantly lower than the loss-only controls by the end of evolution as noted by the non-overlapping 95\% confidence intervals. This gap between the tri-loss-empowerment curve and loss-only control curves lessens for the $k=25$ and $k=45$ time horizons (Figs. \ref{result:shortk}C and \ref{result:shortk}D). Henceforth we focus on only the $k=1$ and $k=45$ tri-loss-empowerment runs in an attempt to uncover the most significant differences between short-term and long-term empowered NCA.

\begin{figure}[htbp]
  \centering
  \includegraphics[width=\linewidth]{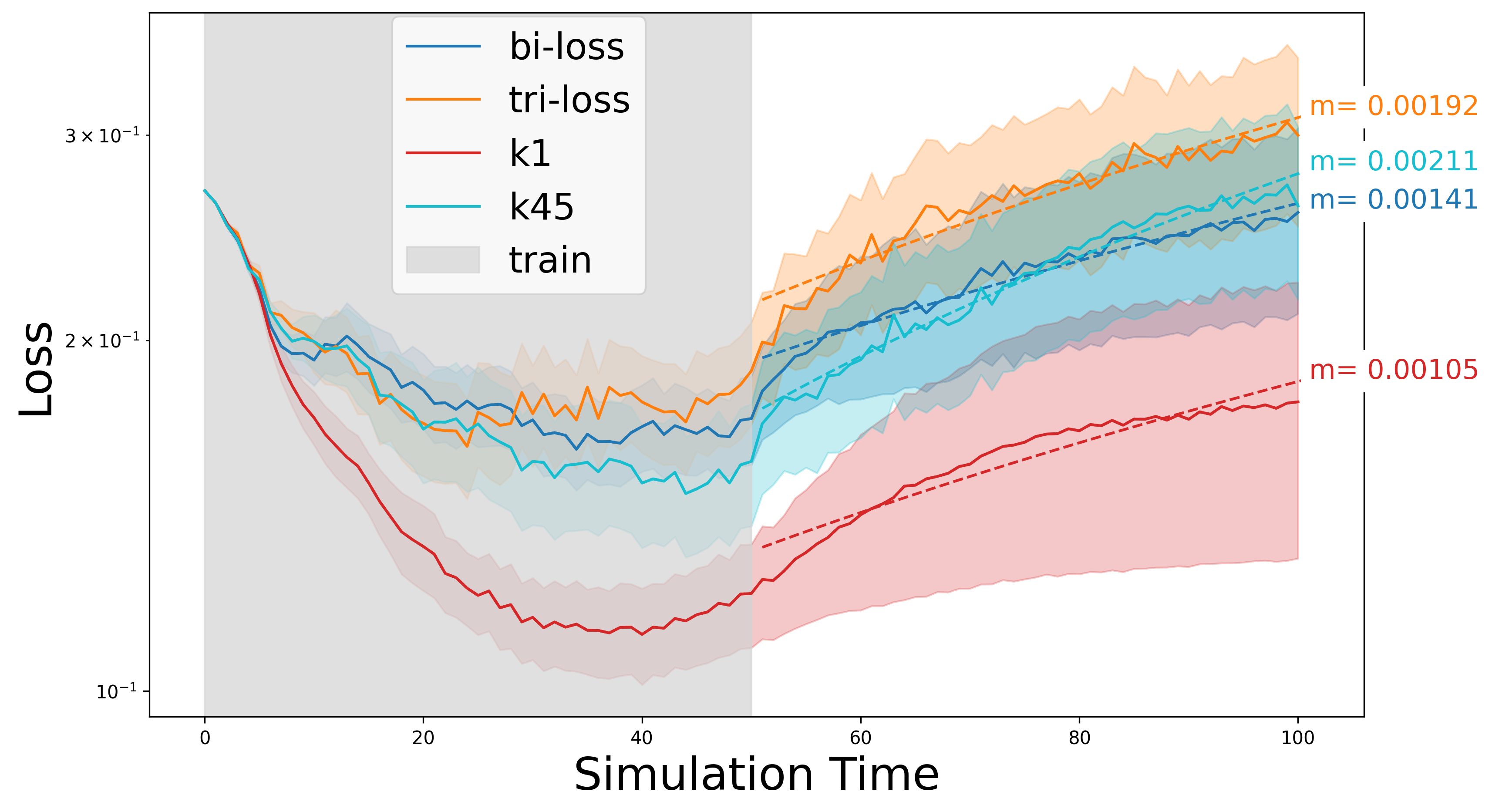}
  \caption{Average loss at time $n$ during NCA simulation of the most fit NCA from each evolutionary trial with 95\% confidence intervals. NCA were run 50 iterations past the simulation time-length during evolution to assess long-term stability. Linear regressions were performed on the average loss curves with the slope $m$ displayed to the right.}
  \Description{Loss curves during NCA simulation for the loss-only treatments and long-term and short-term empowerment trials.}
  \label{results:stability}
\end{figure}

\subsection{Homeostasis}
The most fit NCA at the end of evolution were further investigated to understand defining characteristics of short-term and long-term empowered NCA. One such behavior displayed by short-term empowered NCA is long-term stability. Figure \ref{results:stability} displays loss over developmental time between the NCA's state at iteration $n$ and the target shape. During evolution, NCA are allowed to develop for $N=50$ time-steps before the simulation ends and loss is assessed. This is denoted by the grey shaded region in Figure \ref{results:stability}. We then allowed the NCA to continue updating for 50 additional iterations that were not seen during the evolutionary process. All curves at the end of this additional simulation period trend upward indicating a certain lack of stability, however, the $k=1$ short-term empowered curve is noticeably less steep, noted by the slope of the regression line. This suggests that short-term empowerment produces NCA that are more stable and homeostatic in the sense that they are better able to maintain their form in unseen circumstances.

\begin{figure}[htbp]
  \centering
  \includegraphics[width=\linewidth]{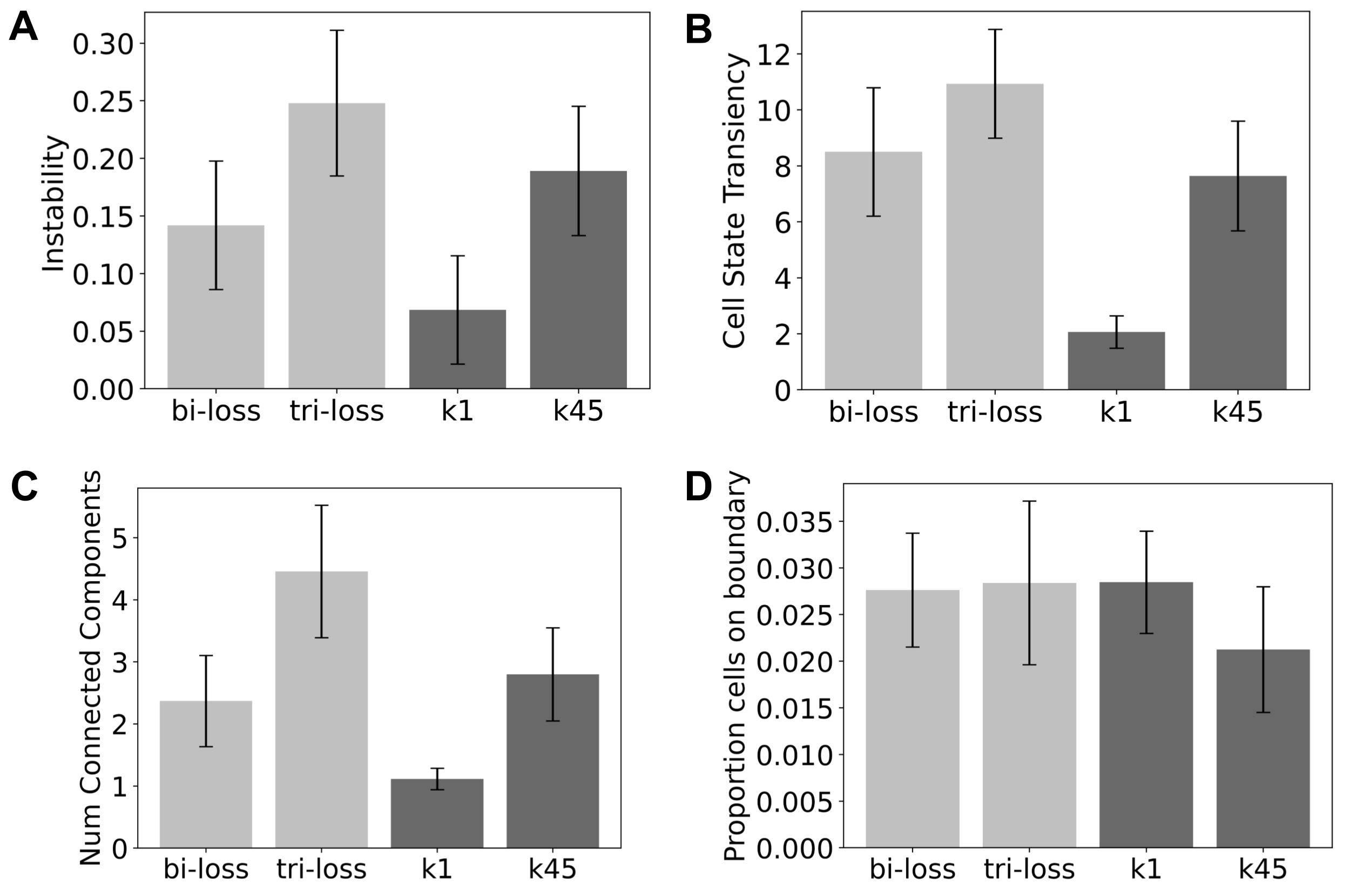}
  \caption{NCA evolved for short-term ($k=1$) empowerment display behaviors that are favorable for the shape-matching task including low long-term instability (A),  low cell transiency (B), and cohesive shapes consisting of few connected components (C) compared to loss-only controls and long-term ($k=45$) empowered NCA. There is, however, no significant difference in proportion of cells that run up against the boundary at the end of simulation. Error bars indicate 95\% confidence intervals.}
  \Description{Four bar plots indicating performance of loss-only controls and the short-term and long-term empowered NCA.}
  \label{results:characteristics}
\end{figure}

This characteristic is further evidenced by another metric used to assess instability as reported in Figure \ref{results:characteristics}A. Here, instability is quantified by computing loss between the final state of the CA after 100 iterations (first 50 time steps seen during evolution, last 50 time steps unseen) and the state of the CA after the first 50 time steps.  
Thus, we observe the NCA only in terms of homeostasis and not morphogenesis as the target shape is not considered here. The short-term empowered NCA have the lowest loss after the additional iterations compared to the shape they initially developed suggesting again that coordination on a short time-scale produces stable, homeostatic behaviors which are also observed in systems capable of morphogenesis. Other characteristics short-term empowered NCA display include low cell transiency, measured by number of times a cell switches states after it initially becomes `alive', and a low number of connected components in the final developed state (Fig. \ref{results:characteristics}B and \ref{results:characteristics}C). To ensure these characteristics were not observed from a perverse behavior such as using the boundary as a means to stop growth, the proportion of cells touching the grid boundary at the end of simulation was measured. We found no significant difference between all variants (Fig. \ref{results:characteristics}D). Taken together, these characteristics suggest that short-term empowered NCA are more likely to produce cohesive and stable shapes, both important qualities of a collective system attempting to match and maintain a target shape.

\begin{figure}[htbp]
  \centering
  \includegraphics[width=\linewidth]{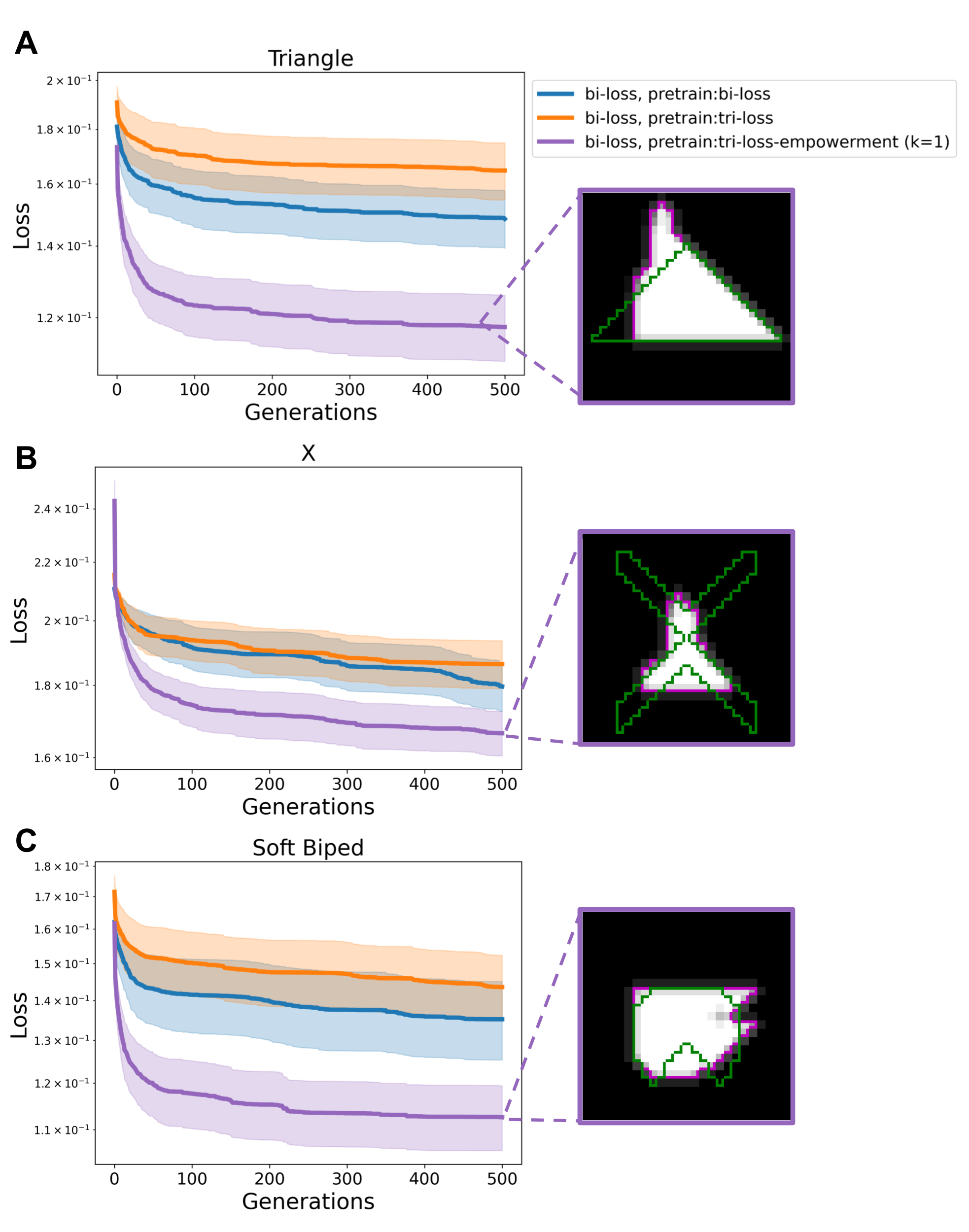}
  \caption{Evolved NCA from the loss-only trials (blue and orange curves) or from the short-term empowered trials (purple and green curves) were used to seed initial populations of evolutionary runs for three different target shapes: a triangle (A), an X (B), and a soft biped shape (C). The blue, orange, and purple curves (evolution seeded with short-term empowered NCA) are evolved for bi-loss and the green curve is evolved for tri-loss-empowerment with $k=1$. The best NCA from the bi-loss runs seeded with short-term empowered NCA are displayed to the right where the green outline indicates the target shape, magenta indicates the shape of the NCA at the end of development, and the grayscale represents the signaling channel.}
  \Description{Loss curves for NCA evolved for 3 different target shapes. The best NCA from the bi-loss runs seeded with short-term empowered NCA are shown to the right of each plot. }
  \label{results:generalization}
\end{figure}

\subsection{Generalization}
Given that the traits short-term empowered NCA exhibit could be considered useful to the task of morphogenesis in general (i.e. not dependent on target shape), we hypothesized that short-term empowered NCA might be more easily tunable to adapt to different target shapes. To this end, the most fit NCA from the loss-only controls and $k=1$ tri-loss-empowerment trials evolved on the square target were used to seed initial populations of three different evolutionary variants. These three variants employed the bi-loss method with the only difference between them being the initial population. Initial populations were seeded with the best NCA pre-trained on the square target with: (1) the bi-loss variant, (2) the tri-loss variant, or (3) the $k=1$ tri-loss-empowerment variant. 
This was repeated for three different target shapes: a triangle, an x-shape, and a soft biped shape. Bi-loss evolutionary runs seeded with short-term empowered NCA (purple curves in Figs. \ref{results:generalization}A, \ref{results:generalization}B, and \ref{results:generalization}C) are lower than the bi-loss runs seeded with non-empowered NCA for each of the three different target shapes tested. It should also be noted that this gap in loss was achieved with a low number of generations (500) compared to the initial runs used to produce the seed populations (2000). Results suggest that empowered NCA exist in a space where it may be easier to reach solutions to different, but similar tasks. The final state of the most-fit NCA after these additional fine-tuning generations for each of the different shapes is displayed to the right of it's respective loss plot in Figure \ref{results:generalization}. The triangle shape appears to be the easiest of the three tasks which is also reflected in the bigger gap between loss of the empowered initialization vs. non-empowered. 

\section{Discussion}

The results above demonstrate that short-term empowerment is more beneficial as an additional objective to loss in NCA attempting the task of shape-matching compared to long-term empowerment as, on average, it produces solutions that are significantly better than evolving shape-matching alone. Various values for time horizon were tested between $k=1$ and the total simulation length, $N=50$, finding that there is a gradual increase in loss between the shortest time horizon $k=1$ and the longest time horizon $k=45$. A simple explanation for this trend could be that as the time horizon increases less of the sensorimotor information from simulation can be used (Fig. \ref{fig:methods}F and \ref{fig:methods}G). A time horizon of $k=45$, for example, has only five sensor/action pairs from which to compute empowerment compared to 49 for a time horizon of $k=1$. Thus, long-term empowerment puts selection pressure on only a few of the NCA's action and sensor states which may inherently make this a more difficult problem.

\begin{figure}[htbp]
  \centering
  \includegraphics[width=\linewidth]{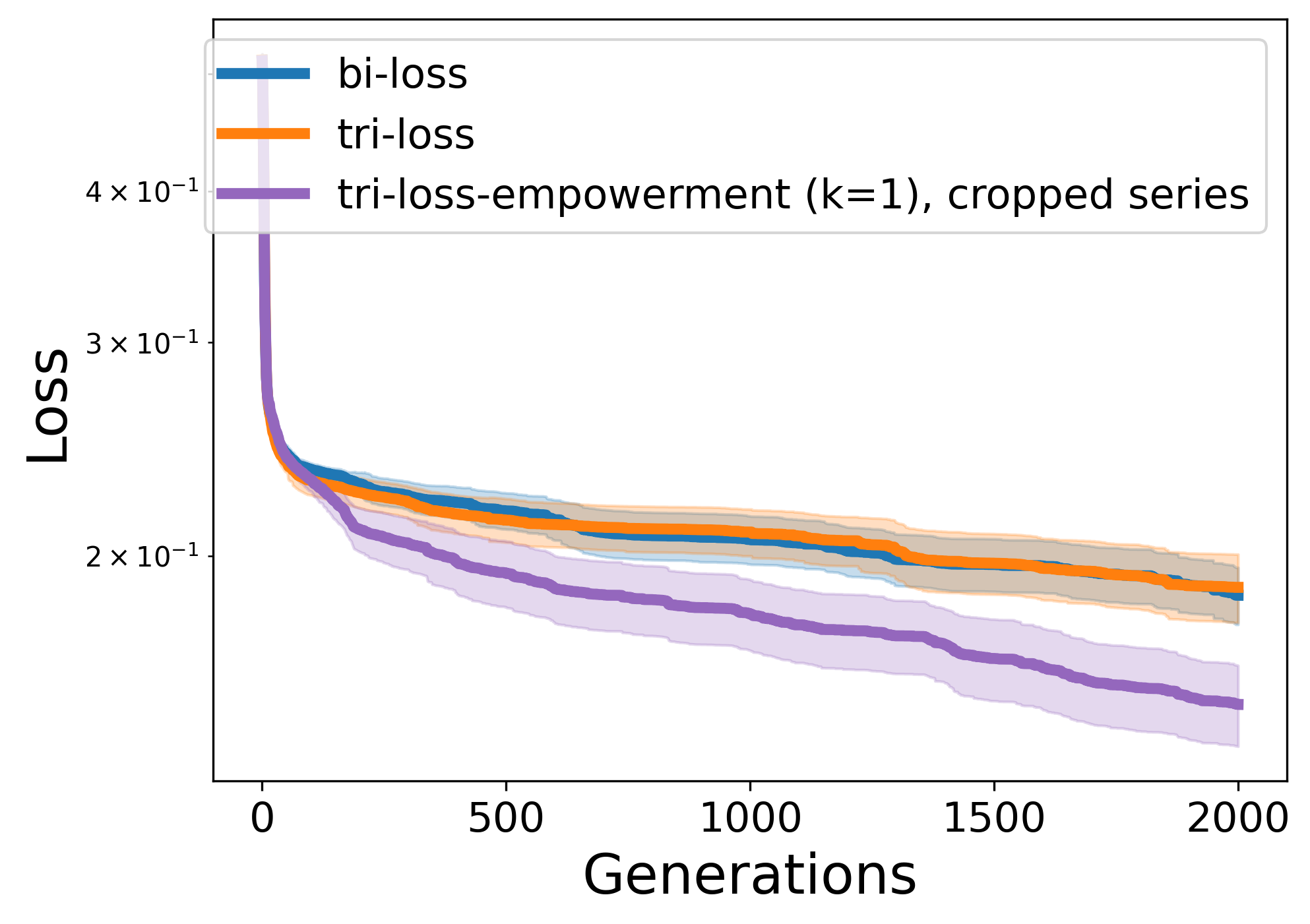}
\caption{Average loss with 95\% confidence intervals of the most fit NCA in the population at each generation. The empowerment computation is done with a time horizon of $k=1$, during which sensor and action time series are cropped to comprise only the last five pairs. This is equal to the times series length of the longest time horizon tested, $k=45$.}
  \Description{Loss plot for for $k=1$ empowerment with a cropped time-series (green curve). The green curve is lower than the loss-only controls which slightly overlapping 95\% confidence intervals.}
  \label{discussion:tslength}
\end{figure}

\begin{figure}[htbp]
  \centering
  \includegraphics[width=0.85\linewidth]{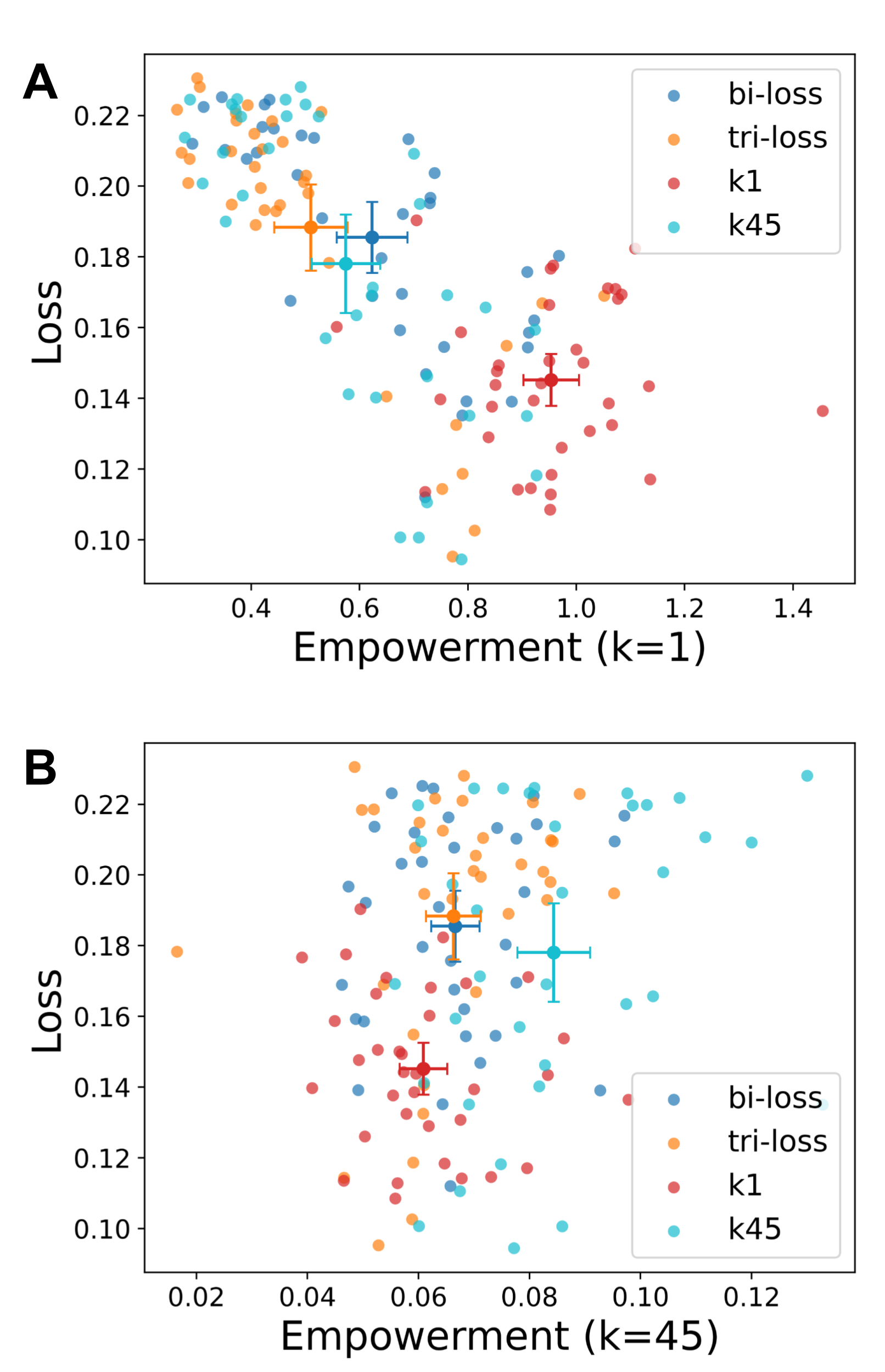} \caption{ The most fit individuals from bi-loss, tri-loss, $k=1$ tri-loss-empowerment, and $k=45$ tri-loss-empowerment trials are plotted in the loss vs. $k=1$ empowerment space (A) and the loss vs. $k=45$ empowerment space (B). Error bars indicate mean and 95\% confidence intervals in each dimension.}
  \Description{Scatter plots of the best solutions from each of the trials in the loss vs. empowerment space for $k=1$ empowerment (A) and $k=45$ empowerment (B).}
  \label{discussion:loss_emp}
\end{figure}

To test this intuition, the experiment for $k=1$ empowerment was repeated, however, during evaluation, the $k=1$ time series were cropped to use only the last five samples for a fair comparison to the $k=45$ runs. Resulting loss over evolutionary time for these runs is showed in Figure \ref{discussion:tslength}. Not surprisingly, we observe that the gap between the $k=1$ tri-loss-empowerment loss curve and the control loss curves has decreased compared to Figure \ref{result:shortk}B where the full time-series was used. However, even with the alteration, short-term empowerment still produces a significant benefit over the loss-only runs compared to that of long-term empowerment (Fig. \ref{result:shortk}D). That is, the benefit of the empowerment on short time horizons is not simply an artifact of the length of the time series using in the empowerment calculation. This suggests that coordination in the NCA's signaling channel on a short time-scale is either more beneficial in this task environment or it is simply a more easily producible behavior in NCA than coordinating signals over long time periods. Evidence to support the latter is displayed in Figure \ref{discussion:loss_emp} which shows the most fit solutions from the loss-only controls and the short-term (without cropped time series) and long-term empowered runs in the loss vs. empowerment space. Figure \ref{discussion:loss_emp}A uses a time horizon of $k=1$ to compute empowerment of the solutions. We observe that the short-term empowered NCA have significantly greater $k=1$ empowerment evidenced by the gap in horizontal confidence intervals. 

In Figure \ref{discussion:loss_emp}B, we visualize the same data but compute $k=45$ empowerment on the solutions. Here, the long-term empowerment NCA don't have as much separation in the x-axis compared to the other solutions, indicating that evolution had more difficulty making progress on this objective. Though such a behavior may still be useful, it is likely difficult for NCA to produce as they operate based on local rules given only information from the previous time-step. As we observe in Figures \ref{result:shortk} and \ref{results:upscale}, medium time horizons also aid in finding more capable solutions, however, further investigation has to be done to understand the intricacies of the relationship between time horizon, total observation length of the system, and task complexity. Without this knowledge, however, it is clear that short-term empowerment consistently outperforms loss alone and thus, could be considered a safe choice for time horizon for empowered NCA.   

Short-term empowered NCA also were shown to display many desirable characteristics of collective systems capable of morphogenesis and homeostasis, such as long-term stability (maintaining form after development), cohesive structures (few holes), and low cell transiency (cells don't flicker or move across the grid). Short-term empowerment as an additional objective pushes solutions toward this behavior without having to directly select for each of these characteristics individually. This allows for easier discovery of fit NCA capable of matching a specified shape in a manner similar to biological organisms where cohesive shapes grow from a single cell and stop when a target shape has been achieved. This benefit of short-term empowerment, thus, produces solutions that have acquired behaviors of morphogenetic systems in general. Indeed, we find that fine-tuning short-term empowered NCA with the bi-loss objectives for a different target shape produce more fit NCA on the new target than fine-tuning non-empowered NCA with the bi-loss objectives for only a short number of generations. This suggests short-term empowered solutions are more easily adapted to different, but similar tasks. This may be beneficial if a new task is desired after already performing evolution on an initial task as it may not be necessary to start from scratch in order to achieve the behavior. Such a finding is a step towards understanding the universal principles that guide processes such as morphogenesis regardless of the specific task (target shape) at hand. 

\begin{figure}[htbp]
  \centering
  \includegraphics[width=0.85\linewidth]{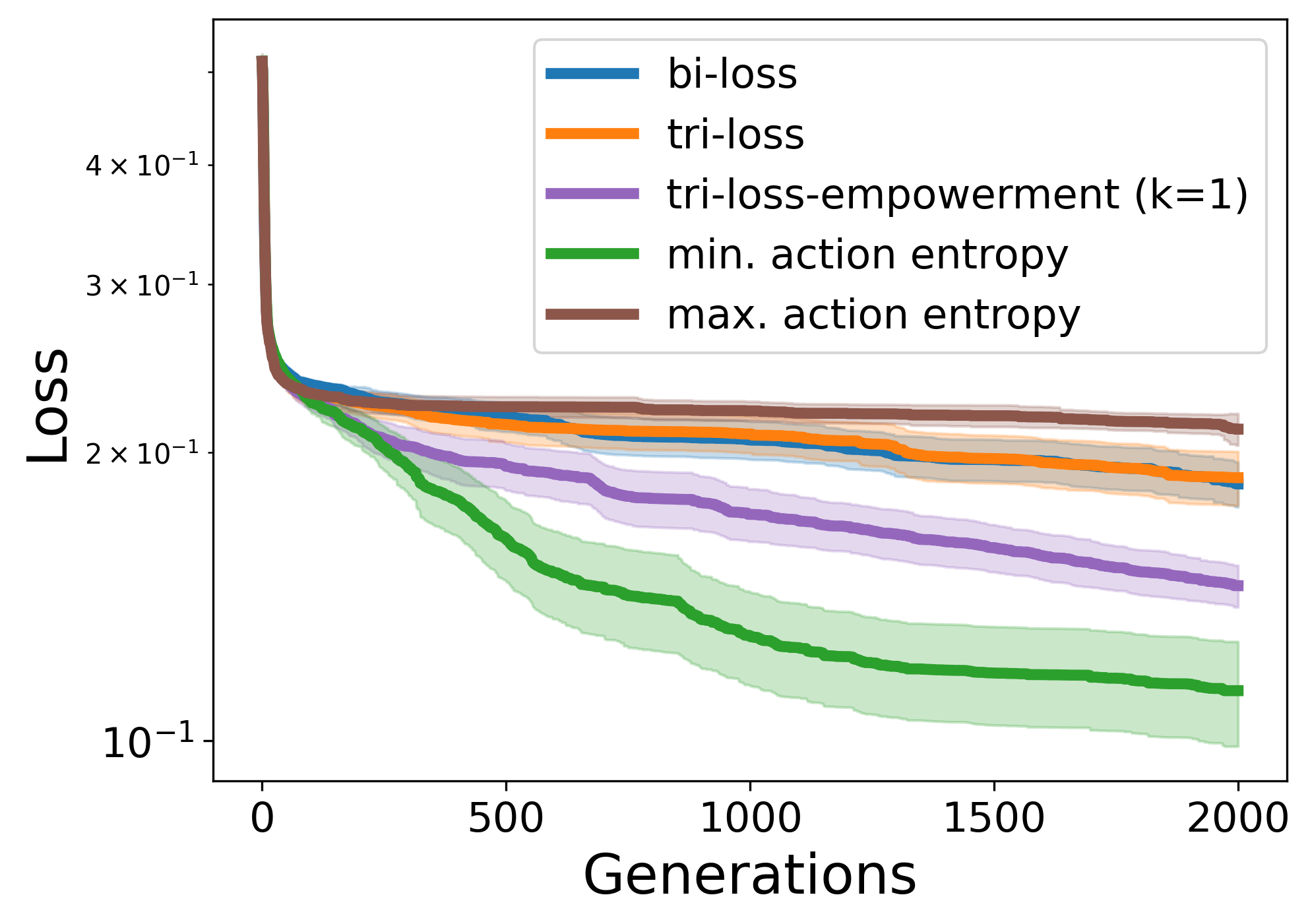} \caption{Loss over evolutionary time averaged over 35 replicates (with 95\% confidence intervals) for the bi-loss, tri-loss, tri-loss-empowerment ($k=1$) and three additional variants: minimizing action entropy of individual cells (local), maximizing entropy of individual cells (local), and minimizing action entropy over all cells (global).}
  \Description{Loss plots showing the additional evolutionary variants discussed in the text.}
  \label{discussion:add_obj}
\end{figure}

Lastly, in an attempt to better understand the types of signaling dynamics that empowerment selects for, we tested alternative additional objectives to replace empowerment with to investigate whether or not they had a similar effect. These included minimizing or maximizing the entropy of individual cells' actions. We found that minimizing the entropy each cell's set of actions significantly outperforms short-term empowerment as an additional objective (Figure \ref{discussion:add_obj}). That is, it is helpful for most cells produce constant actions. 
Empowerment, however, does not appear to have this effect as individual cells tend to explore more of the possible action space. More investigation must be done to determine if empowerment and minimizing local action entropy are indeed selecting for different dynamics, in which case they may be beneficial together, or if there is overlap between the two, in which case we may be able to better understand why empowerment is useful in this task-environment. 

\section{Conclusions and Future Work}

Previously, we found that evolving for loss \textit{and} empowerment of NCA attempting to match and maintain a target shape helps to find more capable NCA compared to evolving for loss alone. We hypothesized, here, that different time scales upon which empowerment operates could impact this observed synergistic relationship between empowerment and morphogenesis. Here, we find that evolving for empowerment with short time horizons relative to the total observation period of the NCA amplifies the synergy. It is not clear, however, whether longer time horizons don't perform as well due to either: (a) a lack of corresponding action, sensor samples with which to compute empowerment or (b) the behavior being more difficult to produce or (c) there is simply something more useful about coordination on short time scales for this particular task. Future work will explore differences in optimal time scales for various tasks other than morphogenesis/homeostasis, such as matching a dynamic pattern or self-organizing into various shapes from multiple seeds as well as better understanding the characteristics of and conditions that result in empowered NCA.


Furthermore, we found that short-term empowered and capable NCA display behaviors that are useful for morphogenesis and homeostasis in general including long-term stability and cohesive development; such characteristics enabled them to be more easily fine-tuned to produce different targets. Future work will include exploring various evolutionary frameworks in which it might be beneficial to pre-train NCA to be empowered followed by fine-tuning for various different tasks.



\begin{acks}
This material is based upon work supported by the National Science Foundation Graduate Research Fellowship Program under Grant No. 1842491. Any opinions, findings, and conclusions or recommendations expressed in this material are those of the author(s) and do not necessarily reflect the views of the National Science Foundation. We also thank the Vermont Advanced Computing Core for providing the computational resources for this work.
\end{acks}

\bibliographystyle{ACM-Reference-Format}
\bibliography{references}

\end{document}